\documentclass{article}

\PassOptionsToPackage{numbers,sort&compress}{natbib}

\usepackage[preprint]{neurips_2026}

\usepackage[utf8]{inputenc}
\usepackage[T1]{fontenc}

\usepackage{hyperref}
\usepackage{url}

\usepackage{booktabs}
\usepackage{makecell}
\usepackage{amsfonts}
\usepackage{nicefrac}
\usepackage{microtype}
\usepackage{xcolor}
\usepackage{graphicx}
\usepackage{amsmath,amssymb}
\newcommand{\CaptionFigOne}{%
\textbf{The chain, and the frozen decoder on unseen mechanisms.}
(a) The six conditions one decision passes through; the mark after evidence use records the failed
transfer gate of Section~5.1, whose cause is inputs outside the decoder's training range. (b) Accuracy of
the frozen decoder on three unseen mechanisms, split by evidence stratum. The 30 full-trace decisions,
from which cheaper evidence was removed by design, score 0.5 because the decoder returns one answer on
balanced labels. The aggregate, 0.84375, is the row-weighted mean of the two groups. The hatched bar is a
linear model fitted after the results, whose registered label-shuffle control failed
(Appendix~\ref{app:diagnostic}).}

\newcommand{\CaptionFigTwo}{%
\textbf{The frozen record.}
One row per setting, one column per condition of Figure~\ref{fig:chain}(a). The unseen-mechanism row's
evidence-use mark records the failed transfer gate of Section~5.1. Glyph shape carries the status, so the
panel survives greyscale. Grey marks a condition that was not tested or not evaluated, and records the
absence of a result, not negative evidence.}

\hypersetup{colorlinks=true,linkcolor=blue!55!black,citecolor=blue!55!black,urlcolor=blue!55!black,
            pdfauthor={Zhengshu Zhang},pdfsubject={}}

\title{When Does Test-Time Physical Diagnosis Pay?}

\author{Zhengshu Zhang\\University of Southern California\\\texttt{zhengshu@usc.edu}}

\begin{document}
\raggedbottom
\maketitle

\begin{abstract}
When a robot faces unfamiliar physical conditions, a common approach is to collect evidence about
what changed and adapt. For such diagnosis to improve behavior, six ordered empirical conditions must
hold: a meaningful reference, identifiability of the physical condition, use of the acquired evidence,
decision value, selection value over a fixed alternative, and safe realization. We test this chain in
controlled and public environments. In our controlled environments the frozen policy chooses correctly
on every decision, but which evidence it buys is fixed by how the panels assign decisions to evidence
strata. On three mechanisms absent from training, the frozen decoder fails its registered transfer
gate. On all 96 decisions of the transfer panel, its full-trace input lies far outside its training
range, and on such inputs it returns one answer whatever the trace contains. All 15 of its errors fall
in the 30 decisions from which cheaper evidence was removed by design, a split hidden by aggregate
accuracy. None of three
primary public tasks passes its registered qualification gates. Before attributing a failure to how a
frozen model uses evidence, evaluation should therefore check whether its deployment inputs lie inside
its training range.
\end{abstract}

\section{Introduction}

Policies are often deployed under physical conditions not seen during training, without retraining or
 data from the new setting \citep{assran2025vjepa2,bobrin2026zeroshot,roeder2026dali}. A common
 response is to infer hidden physical conditions from interaction and use the resulting estimate for
 downstream control or policy selection
 \citep{aoyama2025,killian2017,liang2020,memmel2024,nakhaei2026contextual,raileanu2020,roeder2026dali,zintgraf2020}.
 For this approach to improve behavior, two questions matter: does newly acquired evidence change the
 deployed decision, and does that change improve behavior?

Answering them requires following an adaptation as it actually unfolds. Any claim of improvement first
 needs a reference worth comparing against; otherwise, the gain may simply reflect a weak baseline. Even
 with a strong reference, the available evidence must contain enough information to recover the relevant
 physical condition---identifiability. But information being present does not mean it is being used: a frozen
 decoder may face informative evidence and still keep the same choice, so evidence use must be tested
 separately. If the choice changes, the change may still carry little decision value because the alternatives
 lead to similar outcomes. Even when contexts genuinely favor different choices, a single fixed intervention
 may capture most of the available gain, leaving little selection value for context-dependent choice. A useful
 choice must also remain within the allowed action bounds to be safely realized. These distinctions yield six
 ordered empirical conditions: reference, identifiability, evidence use, decision value, selection value, and
 safe realization. The closest transfer and adapter studies infer the current physical condition themselves
\citep{bobrin2026zeroshot,li2026adapter}. But neither tests whether a frozen decoder uses evidence gathered by the
system itself.

In our controlled environments the frozen policy chooses correctly on every decision, but which
 evidence it buys is fixed by how the panels assign decisions to evidence strata. On three mechanisms
 absent from training, the frozen decoder fails its registered transfer gate. On all 96 decisions of
 the transfer panel, its full-trace input lies far outside its training range, and on such inputs it
 returns one answer whatever the trace contains. All 15 of its errors fall in the 30 decisions from which cheaper evidence was removed
 by design, a split that aggregate accuracy does not reveal. On three public Meta-World tasks, none
 passes the registered qualification gates.
 Identifying the physical condition is only an upstream success; adaptation may still fail before
 behavior improves.

\section{Related Work}

Every component of this audit has a precedent; what we do differently is test conditions apart that
are usually tested together. One line asks whether more evidence should be bought. Sequential sensing and active feature
acquisition study when to keep collecting under cost, with variants that transfer an acquisition
policy to unseen tasks and that evaluate retrospectively when acquisition shifts the distribution;
selective prediction handles when to abstain
\citep{trapeznikov2013,richman2015,valancius2024,dinh2026nocta,aronsson2026afasurvey,schuetz2026afabench,kobayashi2026l2m,vonkleist2025afape,geifman2019,bates2021}.
That line prices evidence but does not test whether a frozen decoder still works on the evidence it
buys once the mechanisms change.

A second line asks how the hidden condition is estimated and what is done with the estimate:
exciting the system so physical quantities can be identified for downstream control, inferring
latent dynamics through hidden parameters or contextual world models, and selecting a policy after
probing
\citep{liang2020,memmel2024,aoyama2025,killian2017,zintgraf2020,nakhaei2026contextual,roeder2026dali,raileanu2020}.
Most of it assumes the context estimate is in hand. The closest transfer result shows a pretrained
behavioral representation failing under dynamics shift with context supplied free
\citep{bobrin2026zeroshot}; that failure is not our finding, and what we add is the price. A third line asks whether it is worth it: whether a measurement pays for itself downstream, that
predictive accuracy is not decision quality, that a correction must stay bounded, and whether
choosing per context beats the single best intervention
\citep{soleymani2021,liu2026dfl,staessens2021,librunskill2026}. Those are our decision-value,
selection-value and safe-realization conditions.

The nearest audit separates same-state headroom, fixed-reference gain and allocation gain for
learned command adapters \citep{li2026adapter}. It too breaks whether an adapter is worth it into
ordered layers, judges them against thresholds registered in advance, and reaches a negative
conclusion. The key difference is that its selector is refit, not frozen, so it does not test a frozen decoder
on unseen mechanisms. A last line is evaluation
methodology itself: statistical practice, task suites, generalization taxonomies, real-to-sim and
crowd-sourced evaluation, benchmark-validity auditing, and off-dynamics benchmarks
\citep{kressgazit2024bestpractices,gao2026taxonomy,wang2026roboeval,jain2025polaris,atreya2025roboarena,liao2026activeeval,yang2025simtorealeval,jiang2026benchmarking,odrl2024}.
Task success rate is traditional there and its coarseness is
already recognised: one study supplements it with completion, sub-task and trajectory measures and a
taxonomy of execution failures \citep{yang2025simtorealeval}, and an evaluation suite reports failure
location and cause alongside performance \citep{robolab2026}. Those decompose an execution that went
wrong. We ask a prior question, about the decision rather than the execution: whether the evidence a
system paid for was used at all, and whether using it would have been worth anything. \section{What Must Hold for Physical Diagnosis to Pay?}

\begin{figure}[!t]
\centering
\includegraphics[width=\textwidth]{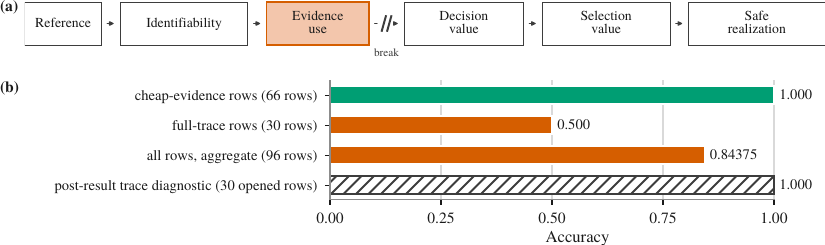}
\caption{\CaptionFigOne}
\label{fig:chain}
\end{figure}

Diagnosis pays only if a chain of conditions holds end to end. Between ``the system found out what
changed'' and ``the system did better for knowing it'' sit six conditions, ordered as a single
decision passes through them (Figure~\ref{fig:chain}a); a break at any link ends the case for
diagnosis however well the earlier ones went. A failed condition ends the case for diagnosis and blocks downstream
development, although diagnostics already available from the same panel may expose later failures.

\textbf{Reference:} the unadapted system must reach a standard set in advance before it can serve as
a comparison. Compare against a weak baseline and the comparison measures the baseline's weakness.
\textbf{Identifiability:} the hidden physical context is recoverable from what the system may
observe. \textbf{Evidence use:} the decoder that turns those observations into a choice was fitted
on one family of mechanisms and then frozen; whether the context is recoverable in principle and
whether this decoder recovers it on unseen mechanisms are different questions, and deployment
depends on the second. \textbf{Decision value:} the context must change which action the system
should take. Value belongs not to the information alone but to the information against the bank of
actions on hand, and a bank whose best member never changes with context makes even a perfect
diagnosis worthless. \textbf{Selection value:} choosing per context must beat applying one fixed intervention
everywhere. \textbf{Safe realization:} the action must be executable. Limits on how large a
correction may be are declared before deployment, and an action exceeding them is clipped rather
than refused, so what runs is not what was chosen.

Acquisition and use are distinct conditions: acquiring is worth it only if the decision made
without it would have been worse, and use happens only if the decoder lets what arrives change its
answer. Neither implies the other.

These are ordered empirical prerequisites, not a causal decomposition and not a theorem; a minimal
formal statement is in Appendix~\ref{app:formal}. Figure~\ref{fig:record} records, for each setting, which conditions were
tested and how each test came out.

\section{Evaluation Protocol: Isolating the Failed Link}

Most thresholds below were registered before the run they judge. The exceptions are the bars of the
after-the-fact diagnostics in Section~5.1 and Appendix~\ref{app:diagnostic}, set after those rows were
opened, and the Ant gates, which were coded without a separate registration. A system that chooses how
much evidence to buy can also beat a fixed rule by buying more. The fixed-rule comparators therefore buy the same
number of each evidence level over a panel, though not the same purchase on each decision. Where
intervals are reported, they are bootstrapped over physical groups, and the methods are compared on the
same rows. The comparisons assume that candidate actions start from the same state. A simulator permits
this by saving the state, trying one candidate, restoring the state and trying the next, which a real
robot cannot do. This assumption was not demonstrated for Walker, Quadruped or the controlled
simulators, and on Ant its registered same-state gate failed.

\section{Findings}
\subsection{The Frozen Decoder Returns One Answer on Full-Trace Decisions}

We first check that the evaluation registers success. We built two simulators, one for residual
tracking and one for differential-drive navigation. In each, every decision chooses between two action
strategies, and which wins depends on a physical condition the system cannot see; in residual tracking
this is whether contact slips, whether the actuator has margin or whether the button breaks. Before
acting, the system buys evidence about that condition at one of four prices, paid in bytes serialized:
nothing, a short event record (two bytes in residual tracking, six values in navigation), a telemetry
summary, or the full trace. Across 2,304 choices in four panels, the frozen policy picks correctly every
time, at zero regret. In residual tracking the panels assign each decision to an evidence stratum by its
index, and the level the policy buys is a fixed function of that stratum and the context. Its cost,
about a third of always buying the full trace, is therefore set by the design, and always buying the
full trace is also correct every time. Two comparators that buy the same number of each evidence level
over a panel, but assign the purchases differently, do worse, because they place some decisions below
the level their stratum needs.

The same acquisition policy and decoder, frozen unchanged, then face three mechanisms neither was
fitted on: stick-slip friction, actuation delay and contact compliance. The panel
generates these three mechanisms with a separate closed-form generator, not with the simulator the
policy was built in.

The decoder is right on 81 of 96 choices. The panel assigns each physical group to one of three
evidence strata by its index. In the first stratum the event bits and the full summary are kept. In the
second, the event bits are zeroed and the summary keeps only its duration and two safety-margin fields.
In the third, the event bits are zeroed and the summary keeps only its duration, so the decisive
quantity appears only in the full trace. These strata are a design that removes cheaper evidence, not a
pattern produced by the acquisition policy. Every decision in the first two strata is correct. All 15
errors fall in the 30 decisions of the third stratum, five per mechanism, and on all 30 the decoder
returns the same class at the same confidence, 0.8775. Because the labels of these decisions are
balanced, this constant answer is right on half.

The full trace of the third stratum does carry the decisive quantity, because the generator places it
there. On every one of these 30 decisions, the better candidate is the one whose trace has the larger
minimum, and the two minima have opposite signs. A fixed rule that compares only these minima, with no
learning, is therefore right on all 30, and so is a linear model on the resampled trace samples. Both
show only that which minimum is larger, a fact the generator fixes, can be read from the trace. Both were also
registered after these rows had been opened, and the linear model's registered label-shuffle control
failed (Appendix~\ref{app:diagnostic}). The decoder's own inputs explain why its answer does not change. On
every one of the 96 decisions, its full-trace input lies far outside its training range. After the
decoder's own standardization, trace values reach 28,806 standard deviations, against at most 10.2 on
its training rows, and the kernel value between the input and every support vector is 0. With every
kernel value at 0, the decoder's output reduces to its intercept. On its own training rows, by contrast, replacing a row's trace with that of
a row with the opposite label changes the predicted probability by 0.915 on average, so inside its
training range the decoder does respond to the trace. These are after-the-fact analyses of
already-opened rows. The conclusion at this gate stands: the registered transfer gate fails. The
failure nearly went unseen. Read alone, 0.84 suggests imperfect transfer and invites more training.
Split by stratum, every error falls in the full-trace stratum, where a constant answer is right on half
the decisions because their labels are balanced, while the other two strata are answered correctly
(Figure~\ref{fig:chain}b). The 0.84 is the row-weighted mean of the two groups. This test asks whether
the frozen decoder transfers; whether a decoder that reads the trace would have changed which strategy
was worth picking needs a different test, run on Ant and Walker in Section~5.3. \begin{figure}[!t]
\centering
\includegraphics[width=\textwidth]{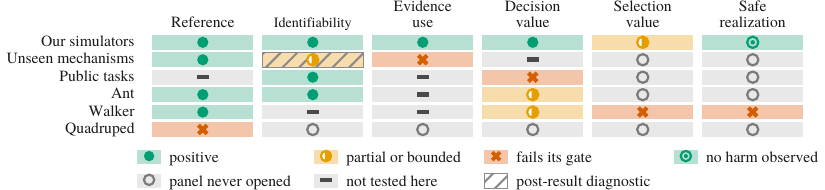}
\caption{\CaptionFigTwo}
\label{fig:record}
\end{figure} \begin{table}[t]
\centering
\small
\caption{\textbf{Public qualification.} Seven registered gates; these are the gates a candidate task
must clear, not the six conditions of Section~3. No primary task passes all seven, and none is
admitted. Ranking reversal, winner diversity and utility separation are computed from the candidates'
rollouts alone. Evidence recovery and early sufficiency are decided by trained probes, and evidence
recovery requires a gain over a no-evidence level that cannot see the context. n/c: recorded as passed,
but not computed from data. The ten-point recovery bar is in Appendix~\ref{app:publicgates}.}
\label{tab:publicgates}
\setlength{\tabcolsep}{3.5pt}
\begin{tabular}{lccccccc l}
\toprule
Task & Protocol & \makecell{Ranking\\reversal} & \makecell{Winner\\diversity} &
\makecell{Utility\\separation} & \makecell{Evidence\\recovery} & \makecell{Early\\sufficiency} &
\makecell{No\\deletion} & Decision \\
\midrule
Door   & pass & fail & pass & pass & fail & pass & n/c & not admitted \\
Drawer & pass & fail & fail & pass & fail & pass & n/c & not admitted \\
Push   & pass & pass & pass & pass & fail & fail & n/c & not admitted \\
Ant    & pass & pass & fail & pass & pass & pass & n/c & replication only \\
\bottomrule
\end{tabular}
\end{table}

\subsection{Public Candidate Problems Fail Before Adaptation Begins}

The controlled result shows that the evaluation registers success when a panel is built to make the
decision solvable. What decides whether to build further is whether any public task needs it. Three were fixed in advance as the primary tasks,
Push, Door and Drawer from Meta-World \citep{yu2020}, and each was checked against the seven registered gates of
Table~\ref{tab:publicgates}. These test that which candidate is better changes with the context, that
different candidates win in different contexts, that the gap between them is worth choosing over, and
that bought evidence improves the decision; the no-deletion gate was recorded without being computed
from data. A task qualifies only if it passes all seven, and development opens only if at least two of
the three primary tasks qualify. None of the three passes (Table~\ref{tab:publicgates}). Door and Drawer already fail the gates
computed from rollouts alone: Door's ordering reverses in one family out of sixteen, and on Drawer the
ordering never reverses and one candidate wins every group. All three also fail the second half of the
recovery gate, which asks that the decision made with the full trace be at least ten points better in
top-1 accuracy than the decision made with no evidence. With the full trace, the probes pick the best
candidate on every Door and Drawer group and on 85.2 percent of the Push groups with a unique winner,
yet the gains over no evidence are 7.1, 0 and 9.3 points. The no-evidence level cannot see the
context, so this comparison is against a fixed prior per family, not against a decision that knows
the context.

Because none of the three passes, development of a selector never opens in the primary chain. Ant
(Gymnasium), a cross-family replication rather than a substitute primary task, fails winner diversity:
one candidate wins 97 of 112 family-context groups, 86.6 percent against a ceiling of 85 percent.
Sixteen of those groups come from a friction context that leaves every candidate's utility unchanged.
Without them the share is 84.4 percent, below the ceiling, so the registered failure depends on counting
a context that changes nothing. Ant's no-evidence level picks the wrong candidate on every group, but
only because its features are identical across candidates and ties go to a candidate that never wins.
A bank whose best member almost never changes is the case Section~3 describes as leaving a perfect
diagnosis worthless, and Section~5.3 measures what the remaining contexts are worth.

Later stages searched further (Appendix~\ref{app:audits}). A single-probe route, which executes one
shared probe, ran all 50 registered attempts, and none passed. A separate development campaign did pass
all seven gates on Push and Door, on two untouched sets of families, and on Ant after an Ant-specific
adapter was developed. Its four candidates, however, are variants of a scripted policy, each run from
the same simulator reset, and its choice reads the first steps of every candidate's own rollout. The
choice is therefore made after every candidate has been executed, which a robot acting once cannot do.
None of these stages shows a public adaptation problem that a robot acting once could exploit, which is
not the same as saying none exists. \subsection{Later Links Fail in Different Ways}

Both failures so far occurred before anything was built on top of the frozen policy. The three
settings below go further. On Ant, on a separate development panel, not the Table~\ref{tab:publicgates}
replication, a 30-step probe raises top-1 accuracy at picking the best candidate from 0.77 to 0.98. Yet
choosing per context improves normalized regret by only 0.00081 over the best fixed candidate, whose
regret is 0.00303. These numbers do not isolate the value of the probe. The 0.77 is a constant
majority-class prediction recorded before the context takes effect, and the gain in accuracy comes from
one context, in which a probe feature directly encodes the change to the applied action. One context
changes nothing, and the run failed its own same-state gate.

Walker (DeepMind Control Suite) fails its qualification at two gates. Each of its policies comes from one
training seed. At 500k steps, a policy trained on default physics averages 974.50 and a robust reference
trained under randomized physics 968.28, over ten evaluation episodes on nominal physics only. On the
60-context panel, the robust reference's mean normalized return is 0.799. Four residual specialists,
one of them for weak actuation, form a five-member bank with the robust reference. To measure how much
better things could be if every choice were right, a selector allowed to see the answer picks the best
member per context across 60 contexts. That selector beats the robust reference by 0.04955 against a
registered floor of 0.05. It is chosen and scored on the same ten-episode means, so this gain is
in-sample. Over the fifteen contexts where the robust reference does worst, the selector gains 0.16.
One residual applied everywhere gains 0.03731, 75.30 percent of the selector's gain, and per-context
choice adds 0.01224 against a registered minimum of 0.02 (Appendix~\ref{app:walkergates}). A
post-result cross-fit chooses members on five episodes and scores them on the other five, over all 252
splits. There the weak-actuation residual is chosen every time, and per-context choice adds 0.0020 on
average. The robust reference has the highest context mean in 14 of the 60 contexts, lies within the
registered tie tolerance of the best in 23, and is the outright winner in 6 of the 30 contexts without
ties.

Realization then fails. The fraction of residual action components clipped at the declared
boundaries runs from 0.19 to 0.22 against a declared maximum of 0.05; clipping was not counted for the
robust reference. The limit was added when the run was locked for execution, after clipping above 0.05
had already been seen. This is a registered action boundary being crossed, not a general claim that
residual actions are harmful, and the action that runs is not the action that was chosen. Walker's
qualification fails because the gates are conjunctive. The selection shortfall is marginal, 0.00045
below its bar with an interval from 0.030 to 0.070, but the clipping violation is not. No fresh
evaluation set was opened for a selector.

Quadruped stops earliest. The policy trained on default physics averages 362.63, 272.22 and 389.49 at
three checkpoints of one training run, below a floor of 400 at each, so the qualification stops at its
first gate, reference competence, which residual training requires. This floor is a supplemental
component added after Walker's rule was transported; the transported relative rule itself passes at
every budget. The averages summarize a bimodal distribution: most episodes return about 190, while 2,
1 and 2 of the ten episodes at the three checkpoints come close to 950. The floor therefore judges a
bimodal policy by an average, as the 81-of-96 aggregate of Section~5.1 averaged perfect scores with a
constant answer; the per-budget returns are in Appendix~\ref{app:quadruped}. A floor added to a
transported rule can end an investigation before adaptation value is measured.

Three settings, three stopping points. On Ant the measured value of choosing per context is
negligible, but the panel does not isolate the probe. Walker falls just short on selection and fails
realization. Quadruped stops at a supplemental floor before any residual is trained. \subsection{An Outcome-Reading Override Repairs; a Broad Override Destroys}

Two rules intervene in decisions: one gains, using information available only after execution, and one
fails badly. On each decision a baseline policy picks one of two candidates, and a rule then keeps that
choice or switches to the other. A switch that makes a wrong baseline right is a correction; one that
makes a right baseline wrong is a destruction. The first rule checks whether the two candidates differ
in a failure bit. This bit records whether a candidate fails the task after its full execution, and the
utility that defines the better candidate includes a penalty for failing, so the rule reads outcome
information that is not available before execution. If the bits differ, the rule settles the decision
on that bit; if not, it leaves the baseline decision unchanged. The rule was designed after
outcomes on the same three mechanisms had been inspected, and its baseline is refit rather than frozen.
The bits differ on 455 of 720 decisions, 63 percent: on all 240 impulse-disturbance decisions, on 215
of 240 rate-limited decisions and on no backlash-hysteresis decision. Over the 720 decisions, accuracy
rises from 0.91 to 0.95, a gain of 0.039 with a group-bootstrap interval of $[0.025, 0.053]$. On every
one of the 455 decisions, the bit-settled choice equals the label, so the rule replaces exactly the
baseline's 28 errors there, and all 28 are corrections. The seeds of these decisions were new, but
their three mechanisms had been used in development. Within its condition the rule therefore acts as a
label oracle, and its gain says nothing about a rule that decides before execution.

An earlier rule built on the same three mechanisms accepted any unanimous, strictly signed agreement
among its source laws, with margin thresholds of 0.0 at every level. Across 1,440 development decisions
it replaced 234, 188 of which turned a correct decision into an error. Per-mechanism figures for both
rules are in Appendix~\ref{app:destructive}. The two results are not comparable. The first rule reads
outcome information, and the second ran on data used for development rather than on sealed evaluation
data. An outcome-reading oracle run on public Ant pair rows also destroyed far more decisions than it
corrected (Appendix~\ref{app:destructive}). \section{Limitations}

All six conditions were screened in simulation, enabling cheap falsification but limiting generality.
Our simulators show only that the sequence can work under conditions we set; public results remain
specific to the candidates, fault families and thresholds. Walker used an answer-seeing selector,
and Quadruped stopped at its first gate. The constant full-trace
answer appeared on every unseen-mechanism decision, with cheaper evidence intact or removed, and the
decoder's inputs there lie far outside its training range (Section~5.1); a decoder trained on wider data
was not tested. No public task passed qualification under a protocol in which the system acts once, and
the one development stage that passed chooses after executing every candidate
(Appendix~\ref{app:audits}), so public occurrence remains open. The transfer result uses one decoder,
one acquisition policy and three mechanisms with one seed each, so it shows that such a failure can
occur, not how common it is (Appendix~\ref{app:scale}). We recommend checking whether deployment inputs
lie inside a frozen model's training range, and reporting results by evidence level when acquisition
depth varies. The six conditions ask different questions and measure different things, so they are independent
tests rather than repetitions of one experiment. Where thresholds were fixed in advance, fixing them blocked adjusting the criterion
after seeing the result, but it also exposed comparator instability such as the bimodal Quadruped returns. Section~5.4
compares only two rules, and the conditioned rule reads outcome information, so its gain does not show
that a rule deciding before execution would help. Neither that result nor the six-condition separation establishes general safety or a universal
causal order, and diagnosis may pay elsewhere.
\section{Conclusion}

No setting in this record passed every condition. In the controlled environments the policy was always
correct, with purchases set by the panel design. On unseen mechanisms the frozen decoder failed its
transfer gate because its inputs lay far outside its training range, and aggregate accuracy hid that
every error fell in one stratum. The primary public tasks failed qualification before any selector was
developed. On Ant the panel could not isolate the value of the probe, Walker fell just short on
selection and failed realization, and Quadruped stopped at a supplemental floor. A single end-to-end
score compresses these outcomes into one number, while the repair each calls for differs. This record
offers no new controller and no universal audit protocol. It offers three practices that can be used
directly. The first is to check whether deployment inputs lie inside a frozen model's training range
before attributing its failure to how it uses evidence. The second is to check any aggregate against a
breakdown by evidence level, and the third is to test the conditions in order, so that a case ends at
the first failed condition.
Diagnosis pays only when the evidence is read after transfer, what is read changes which decision is
best, per-context choice beats the best fixed intervention, and the chosen action stays inside its
boundary. {\small
\bibliographystyle{plainnat}
\bibliography{references}
}

\clearpage
\appendix \section{Minimal formal statement}\label{app:formal}

Write $z$ for the hidden physical context and $\mathcal{A}$ for a finite bank of interventions or
continuations, so that for a value $V(a,z)$ the privileged best member is
\[a^{*}(z) = \arg\max_{a \in \mathcal{A}} V(a,z).\]
A model that predicts $z$, or $a^{*}(z)$, accurately has cleared the second condition and no more;
it has not shown that acting on $a^{*}(z)$ beats sticking with one fixed member of the bank.
Identifiability, decision value, selection value and safe realization form a descending chain.
Nothing in the definitions makes one link imply the next, although this record does not show a
failure for every adjacent pair. Evidence use is tracked separately because it tests evidence
interpretation rather than privileged downstream headroom.

\section{Scale of the headline result}\label{app:scale}

The headline result of Section~5.1 comes from one decoder, one acquisition policy, three mechanisms
and 96 decisions, of which 30 belong to the stratum in which only the full trace carries the decisive
quantity. The diagnostics of Appendix~\ref{app:diagnostic} are smaller still: twenty decisions to train and ten to test in each
fold. These show that the transfer failure can occur, not how common it is.

\section{The post-result tests on the full-trace rows}\label{app:diagnostic}

A fixed rule that compares the two candidates' trace minima, with no learning, is right on all 30
decisions. In this stratum the generator places the decisive quantity in the trace and removes it from
the summary, so this rule shows that the information is present by construction. The rule was
registered after the rows had been opened.

A linear model on the two traces, each resampled to 64 points, is also right on all 30 decisions. It is
fitted leave-one-mechanism-out, with twenty decisions to fit and ten to test in each fold. On these rows
the better candidate is always the one whose trace has the larger minimum, and the two minima have
opposite signs, so a correct linear read-out shows only that which minimum is larger can be read from the samples. A
model given the summary instead scores 0.5, but this control is degenerate: the summary inputs are the
same vector on all 30 rows, so any such model predicts one class and scores 0.5 on balanced labels. The
registered label-shuffle control also failed. Shuffling the labels by physical group 300 times gave null
accuracies from 0.0 to 1.0, with median 0.5 and 95th percentile 1.0, and 49 of the 300 exceeded 0.75.
Each fold's training groups flip as whole units, so the null itself could not meet the registered bar of
at most 0.75 for its 95th percentile.

\section{Public gate detail}\label{app:publicgates}

Drawer's opening observation picks the correct candidate in every context, the same candidate wins
throughout, and the ordering never reverses across families. Door's ordering reverses in one family
out of sixteen and its full-trace gain in top-1 is seven points against a ten-point bar. Push comes
closest, with reversals, diverse winners and pairwise accuracy lifted from 0.71 to 0.95, but a top-1
gain of 0.093 against that same bar and a bootstrap lower bound below zero. Nor could a confidence
bound be calibrated at any cheaper level at which the system could safely stop buying.

\section{Later public stages}\label{app:audits}

Widening the choice from two candidates to four, to see whether the interface still holds: only Door
is fully supported at four. Held-out coverage on Push and Ant falls to 0.67 and 0.58, measured against a
reference attempt chosen after the search, on held-out panels opened earlier. A single-probe route: all 50
registered attempts failed, but 48 of them could not have passed with any selector, because the
candidate bank already failed the rollout-only problem gates on the panel they opened; 47 of them used
the same probe bank. Door put back through
qualification: blocked near its thresholds, with the next set of evaluation data left untouched. Push
put back through qualification and through a search for new candidate formulations: neither passes. A
separate development campaign passed all seven gates on Push and Door on two untouched sets of families,
and on Ant after an Ant-specific adapter was developed. Its choice, however, reads the first steps of
every candidate's own rollout, each run from the same simulator reset, so it is made after every
candidate has been executed.

\section{Quadruped reference returns}\label{app:quadruped}

\begin{figure}[h]
\centering
\includegraphics[width=0.62\textwidth]{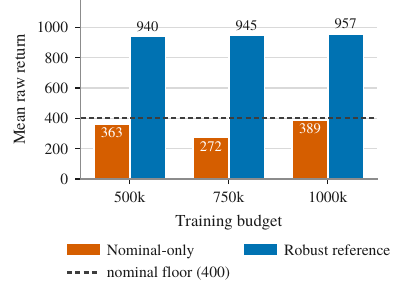}
\caption{\textbf{Quadruped reference qualification.} The policy trained on default physics stays below the
supplemental floor of 400 at every budget, while the robust reference returns 940 to 957, so the
qualification stops at its first gate.}
\label{fig:quadruped}
\end{figure}

At 500k, 750k and 1M steps of one training run, the policy trained on default physics returns 362.63,
272.22 and 389.49 on average, all below the floor of 400. This floor is a supplemental component added
after the transport. It is the only failing component, and the registered relative rule passes at every
budget. The averages hide a bimodal distribution: in most episodes the policy returns about 190, while 2,
1 and 2 of the ten episodes at the three checkpoints come close to 950
(Figure~\ref{fig:quadruped}).

\section{The two destructive records}\label{app:destructive}

The unconditioned rule of Section~5.4 replaced 234 decisions, with 46 corrections and 188 destructions.
On the 11 decisions where the two candidates' failure bits differ, it replaced at the event level and
made 11 corrections. On the 223 where the bits are equal, it replaced after full-trace acquisition and
made 35 corrections and 188 destructions, so these rows cannot separate the effect of full-trace
acquisition from that of equal failure bits. A public diagnostic used a different rule, a prefix ranker
that replaces the baseline on every group, on different tasks and candidates. It made 16 corrections
against 24 destructions on 1,800 Ant pair rows, and 42 against 54 on 1,800 Door pair rows. An
outcome-reading oracle on the same public rows, which settles a pair when the candidates' terminal
statuses differ, made 22 corrections against 221 destructions on Ant and 2 against 0 on Door.

Per mechanism, the conditioned rule takes impulse disturbance from 0.95 to 1.00 and rate-limited
actuation from 0.92 to 0.98, while on backlash hysteresis it changes no outcome at all and leaves
that mechanism level with the baseline. The
unconditioned rule's 188 destructions all fall on decisions with equal failure bits that bought the full
trace, and 172 of them on backlash hysteresis, the one mechanism on which the conditioned rule never
acts.

\section{Walker selection-gate values}\label{app:walkergates}

\begin{figure}[h]
\centering
\includegraphics[width=\textwidth]{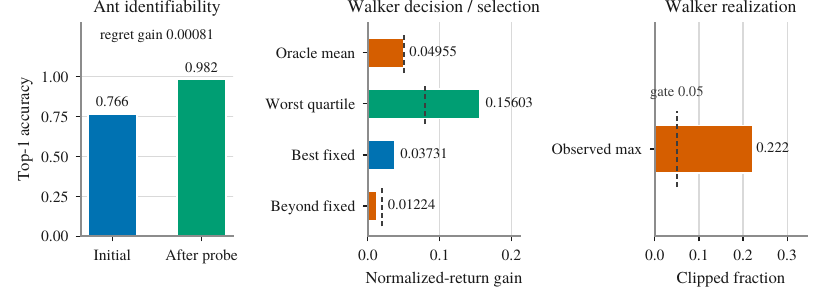}
\caption{\textbf{Ant and Walker gate values.} Left: on Ant, top-1 accuracy rises from 0.766, a
majority-class prediction recorded before the context takes effect, to 0.982 after the probe, while
per-context choice improves regret by 0.00081. Centre: on Walker, in-sample, the selector allowed to
see the answer gains 0.04955 in the mean and 0.15603 on the worst quartile, but one fixed residual already
captures 0.03731 of that and per-context choice adds only 0.01224 beyond it. Right: the largest
clipped fraction, 0.222, against the registered maximum of 0.05. A dashed marker is that bar's own
registered gate; the 0.08 marker on the worst quartile comes from an overlay, not from the frozen
threshold source.}
\label{fig:antwalker}
\end{figure}

The best fixed member of the bank, the weak-actuation residual, gains 0.03731 applied everywhere,
which is 75.30 percent of the 0.04955 the selector allowed to see the answer gains in the mean.
Per-context choice adds 0.01224 beyond that fixed residual, below the registered minimum of 0.02. All
three figures are in-sample. Figure~\ref{fig:antwalker} places these
alongside the Ant result and the clipping fraction.

\end{document}